\def\year{2021}\relax
\documentclass[letterpaper]{article} % DO NOT CHANGE THIS
\usepackage{aaai21}  % DO NOT CHANGE THIS
\usepackage{times}  % DO NOT CHANGE THIS
\usepackage{helvet} % DO NOT CHANGE THIS
\usepackage{courier}  % DO NOT CHANGE THIS
\usepackage[hyphens]{url}  % DO NOT CHANGE THIS
\usepackage{graphicx} % DO NOT CHANGE THIS
\usepackage{natbib}  % DO NOT CHANGE THIS AND DO NOT ADD ANY OPTIONS TO IT
\usepackage{caption} % DO NOT CHANGE THIS AND DO NOT ADD ANY OPTIONS TO IT
\usepackage[switch]{lineno}  %

\usepackage{soul}
\usepackage{tikz}

\newcommand{\squishlist}{
    \begin{list}{$\bullet$}
        { \setlength{\itemsep}{0pt}      \setlength{\parsep}{0pt}
            \setlength{\topsep}{0.5pt}       \setlength{\partopsep}{0pt}
            \setlength{\listparindent}{-2pt}
            \setlength{\itemindent}{-5pt}
            \setlength{\leftmargin}{0.5em} \setlength{\labelwidth}{0em}
            \setlength{\labelsep}{0.2em} } }
    
\newcommand{\squishend}{
\end{list}  }

\newcommand{\COMMENT}[1]{#1}

  \renewcommand{\COMMENT}[1]{}

\newcommand{\tabincell}[2]{\begin{tabular}{@{}#1@{}}#2\end{tabular}}  

\usepackage{threeparttable}
\usepackage{longtable}
\usepackage{makecell}
\usepackage{subfigure}
\usepackage{multirow}
\usepackage{mathtools}
\usepackage{adjustbox}
\usepackage{booktabs}
\usepackage{bm}

\usepackage{amsmath, amssymb,amsfonts}

\title{YOLObile: Real-Time Object Detection on Mobile Devices via Compression-Compilation Co-Design}

\author{
    Yuxuan Cai\textsuperscript{\rm 1}\thanks{These Authors contributed equally.}, 
    Hongjia Li\textsuperscript{\rm 1}\footnotemark[1],
    Geng Yuan\textsuperscript{\rm 1}\footnotemark[1],
    Wei Niu\textsuperscript{\rm 2},
    Yanyu Li\textsuperscript{\rm 1},\\
    Xulong Tang\textsuperscript{\rm 3},
    Bin Ren\textsuperscript{\rm 2},
    Yanzhi Wang\textsuperscript{\rm 1 }
    \\
}
\affiliations{
    \textsuperscript{\rm 1}Northeastern University\\ \textsuperscript{\rm 2}William \& Mary\\ \textsuperscript{\rm 3}University of Pittsburgh\\
    \textsuperscript{\rm 1}\{cai.yuxu, li.hongjia, yuan.geng, li.yanyu, yanz.wang\}@northeastern.edu,
    \\\textsuperscript{\rm 2}wniu@email.wm.edu, 
    \textsuperscript{\rm 2}bren@cs.wm.edu, \textsuperscript{\rm 3}tax6@pitt.edu

}

\begin{document}
% \linenumbers  %

\maketitle

\begin{abstract}
The rapid development and wide utilization of object detection techniques have aroused attention on both accuracy and speed of object detectors. However, the current state-of-the-art object detection works are either accuracy-oriented using a large model but leading to high latency
or speed-oriented using a lightweight model but sacrificing accuracy. In this work, we propose YOLObile framework, a real-time object detection on mobile devices via compression-compilation co-design. A novel block-punched pruning scheme is proposed for any kernel size. To improve computational efficiency on mobile devices, a GPU-CPU collaborative  scheme is adopted along with advanced compiler-assisted optimizations. Experimental results indicate that our pruning scheme achieves 14$\times$ compression rate of YOLOv4 with 49.0 mAP. 
Under our YOLObile framework, we achieve 17 FPS inference speed using GPU on Samsung Galaxy S20. 
By incorporating our proposed GPU-CPU collaborative scheme, the inference speed is increased to 19.1 FPS, and outperforms the original YOLOv4 by 5$\times$ speedup.
Source code is at: \url{https://github.com/nightsnack/YOLObile}.

\end{abstract}

\section{Introduction}
\label{sec:intro}

Object detection, one of the major tasks in the computer vision field, has been drawing extensive research from both academia and industry thanks to the breakthrough of deep neural network (DNN). Object detection is widely adopted in numerous computer vision tasks, including image annotation, event detection, object tracking, segmentation, and activity recognition, with a wide range of applications, such as autonomous driving, UAV obstacle avoidance, robot vision, human-computer interaction, and augmented reality.
Considering these application scenarios, it is equivalently important to maintain high accuracy and low latency simultaneously when deploy such applications on resource-limited platforms, especially mobiles and embedded devices.

In the past decades, promising object detection approaches are proposed, which are mainly categorized into two-stage detectors \cite{girshick2014rich,girshick2015fast,ren2015faster,he2017mask} and one-stage detectors \cite{redmon2016you,redmon2017yolo9000,redmon2018yolov3,bochkovskiy2020yolov4,liu2016ssd,lin2017focal}. Compared with two-stage detectors, one-stage detectors aim to provide an equitable trade-off between accuracy and speed, and will be mainly discussed in this work. Despite large efforts devoted, representative works such as You Only Look Once (YOLO) \cite{redmon2016you,redmon2017yolo9000,redmon2018yolov3,bochkovskiy2020yolov4}, Single Shot Detector (SSD) \cite{liu2016ssd}, still require extensive computation to achieve high mean average precision (mAP), result in the main limitation for real-time deployment on mobile devices. Apart from large-scale approaches mentioned above, lightweight object detection architectures targeted for mobile devices are investigated \cite{sandler2018mobilenetv2,huang2018yolo,li2018tiny}. However, the accomplished efficiency leads to non-negligible accuracy drop.

To address this issue in object detection detectors, model compression techniques have been drawing attention, especially weight pruning methods, which have been proved as one of the most effective approaches to reduce extensive computation and memory intensity without sacrificing accuracy \cite{wen2016learning,guo2016dynamic,min20182pfpce,he2018amc,he2019filter}. By reducing the vast redundancy in the number of weights, models with structural sparsity achieve higher memory and power efficiency and low latency during inference. Generally, unstructured pruning and structured pruning are the two main trendy schemes of weight pruning. Unstructured pruning eliminates weights in an irregular manner, which causes the essential drawback to obstruct hardware accelerations \cite{han2015learning,guo2016dynamic,liu2018rethinking}. Structured pruning is observed with notable accuracy degradation due to the coarse-grained nature in pruning whole filters/channels \cite{min20182pfpce,zhuang2018discrimination,zhu2018improving,ma2019tiny,zhao2019variational,Liu2020Autocompress}. To overcome these shortcomings, pattern-based pruning is proposed incorporating fine-grained unstructured pruning in a hardware aware fashion~\cite{ma2020pconv,niu2020patdnn}. However, it is only applicable to convolutional (CONV) layers with 3$\times$3 kernels, significantly limiting its deployment in object detection tasks. 

The goal of this paper is to achieve real-time object detection by exploiting the full advantages of pruning for inference on mobile platforms.
We propose YOLObile framework, a real-time object detection on mobile devices via compression-compilation co-design. State-of-the-art detectors YOLOv4~\cite{bochkovskiy2020yolov4} is adopted as our detection architecture. YOLObile consists of several novel techniques including a new block based pruning for arbitrary kernel size, compiler optimizations, and a GPU-CPU collaborative acceleration strategy.  
To be more specific, we propose a novel pruning scheme—-block-punched pruning, which is flexible and applicable to CONV layers with \emph{any} kernel size as well as fully-connected (FC) layers. The whole DNN weights from a certain layer are divided into a number of equal-sized blocks and the weights within a block are pruned to a same shape. To improve the computational efficiency of DNNs on mobile devices, the proposed YOLObile also adopts a GPU-CPU collaborative computation scheme. As a result, YOLObile achieves high hardware parallelism using our proposed compiler optimizations, including compact storage scheme, block reordering, and highly parallel auto-tuning model employment.
Experimental results indicate that YOLObile delivers 14$\times$ compression rate (in weights) of YOLOv4 with 49.0 mAP. It achieves 19.1 frames per second (FPS) inference speed on an off-the-shelf Samsung Galaxy S20, and is 5$\times$ faster than the original YOLOv4. 

\section{Background}
\label{sec:background}
\subsection{Preliminaries on Object Detection DNNs}

The DNN-based object detectors can be categorized into two mainstreams: (i) two-stage detectors and (ii) one-stage detectors. 

\textbf{Two-stage detectors} divide the detection to two stages: extract region of interest (RoI) and then do the classification and bounding box regression tasks based on the RoI. A most representative series of two-stage detectors is R-CNN~\cite{girshick2014rich} with its extended generations Fast R-CNN~\cite{girshick2015fast} and Faster R-CNN~\cite{ren2015faster}. R-CNN is the first region-based CNN object detector and it achieves higher object detection performance compared with previous HOG-like features-based systems~\cite{liu2020deep}. Through ensuing development, Faster R-CNN has improved both precision and detection efficiency. Despite highest accuracy rates achieved, the major drawback of such two-stage detectors is high computation and still relatively slower inference speed due to the two-stage detection procedure.

\textbf{One-stage detectors} eliminate the RoI extraction stage and directly classify and regress the candidate anchor boxes.
% OverFeat~\cite{sermanet2013overfeat} is the first one-stage detector.
YOLO \cite{redmon2016you} adopts a unified architecture that extracts feature maps from input images, then it regards the whole feature maps as candidate regions to predict bounding boxes and categories.
YOLOv2~\cite{redmon2017yolo9000}, YOLOv3~\cite{redmon2018yolov3} and YOLOv4~\cite{bochkovskiy2020yolov4} are proposed with improved speed and precision.
SSD~\cite{liu2016ssd} is another representative one-stage detector, establishing a fully convolution network to predict a fixed numbers of bounding boxes and scores. 
One-stage detectors demonstrate an optimized trade-off between accuracy and speed only on high performance desktop GPUs. 

Therefore, lightweight object detectors are proposed for mobile devices where both model size and speed are limited.
SSDLite~\cite{Sandler_2018} is a mobile friendly variant of regular SSD utilizing a mobile architecture, MobileNet, which is based on an inverted residual structure and uses lightweight depthwise convolutions to filter features. Based on YOLOv2, YOLO-LITE~\cite{huang2018yolo} provides a smaller, faster, and more efficient model increasing the accessibility of real-time object detection to a variety of devices. 
Tiny-DSOD~\cite{li2018tiny} is based on a deeply supervised object detection (DSOD) framework and dedicates to resource-restricted usages by adopting depthwise dense block (DDB) based backbone and depthwise feature-pyramid-network (D-FPN) based front-end. 
However, the accuracy of these works is sacrificed significantly. 
On COCO dataset under AP@[0.5:0.95] metric, the accuracy of YOLO-LITE decreases to 12.16 ~\cite{huang2018yolo}, while SSDLite~\cite{Sandler_2018} achieves only 22.1 and Tiny-DSOD obtains 23.2 ~\cite{li2018tiny}.

\subsection{DNN Model Pruning}
\begin{figure}[!t]
    \centering
    \includegraphics[width=0.95\columnwidth]{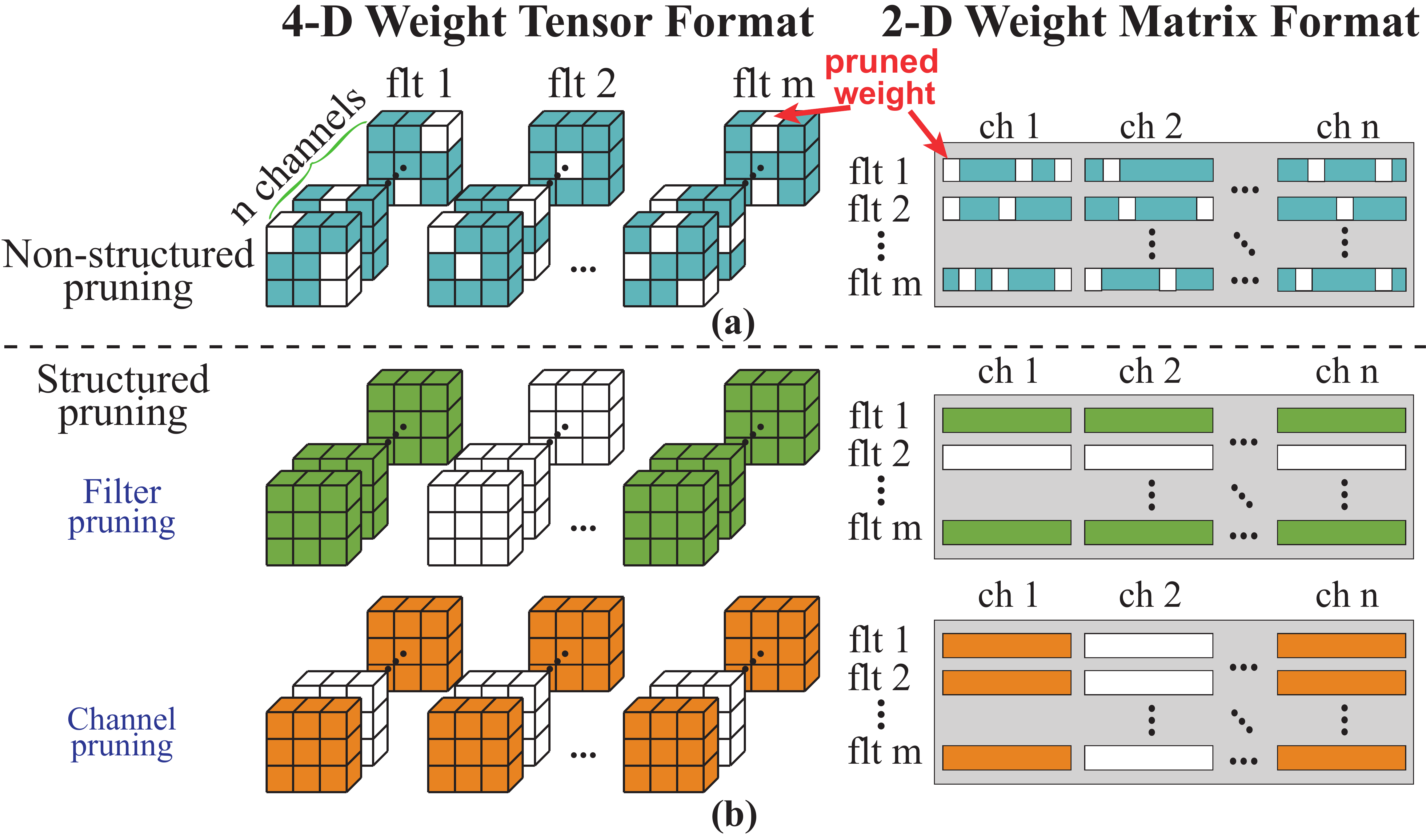}
    \caption{Illustration of (a) unstructured pruning and (b) coarse-grained structured pruning.}
    \label{fig:structured}
\end{figure}

We now discuss the three most trendy pruning schemes: including fine-grained unstructured pruning, coarse-grained structured pruning, and pattern-based pruning.

\textbf{Unstructured pruning} allows the weights at arbitrary locations in the weight matrix to be pruned, which ensures a higher flexibility to search for optimized pruning structure~\cite{guo2016dynamic,frankle2018lottery,dai2019nest}, as shown in Figure~\ref{fig:structured} (a). Thus, it usually achieves high compression rate with minor accuracy loss.
However, unstructured pruning leads to an irregular sparsity in the weight matrix, which requires additional indices to locate the non-zero weights during the computation. This makes the hardware parallelism provided by the underlying system (e.g., GPUs in mobile platforms) underutilized. 
Consequently, the unstructured pruning is not applicable for DNN inference acceleration, and even a decrease in speed can be observed~\cite{wen2016learning}.

\textbf{Structured pruning} prunes the entire channel(s)/filter(s) of DNN weights~\cite{wen2016learning,he2017channel,he2019filter,yu2018nisp}.
As Figure~\ref{fig:structured} (b) shows, the filter pruning removes whole row(s) of the weight matrix, where the channel pruning prunes the consecutive columns of  corresponding channel(s) in the weight matrix. 
Structured pruning maintains the regular shape of the weight matrix with reduced dimension. 
Therefore, it is hardware friendly and can leverage the hardware parallelism to facilitate acceleration.
However, structured pruning suffers from considerable accuracy loss due to its coarse-grained pruning feature.

\begin{figure}[!t]
    \centering
    \includegraphics[width=0.97\columnwidth]{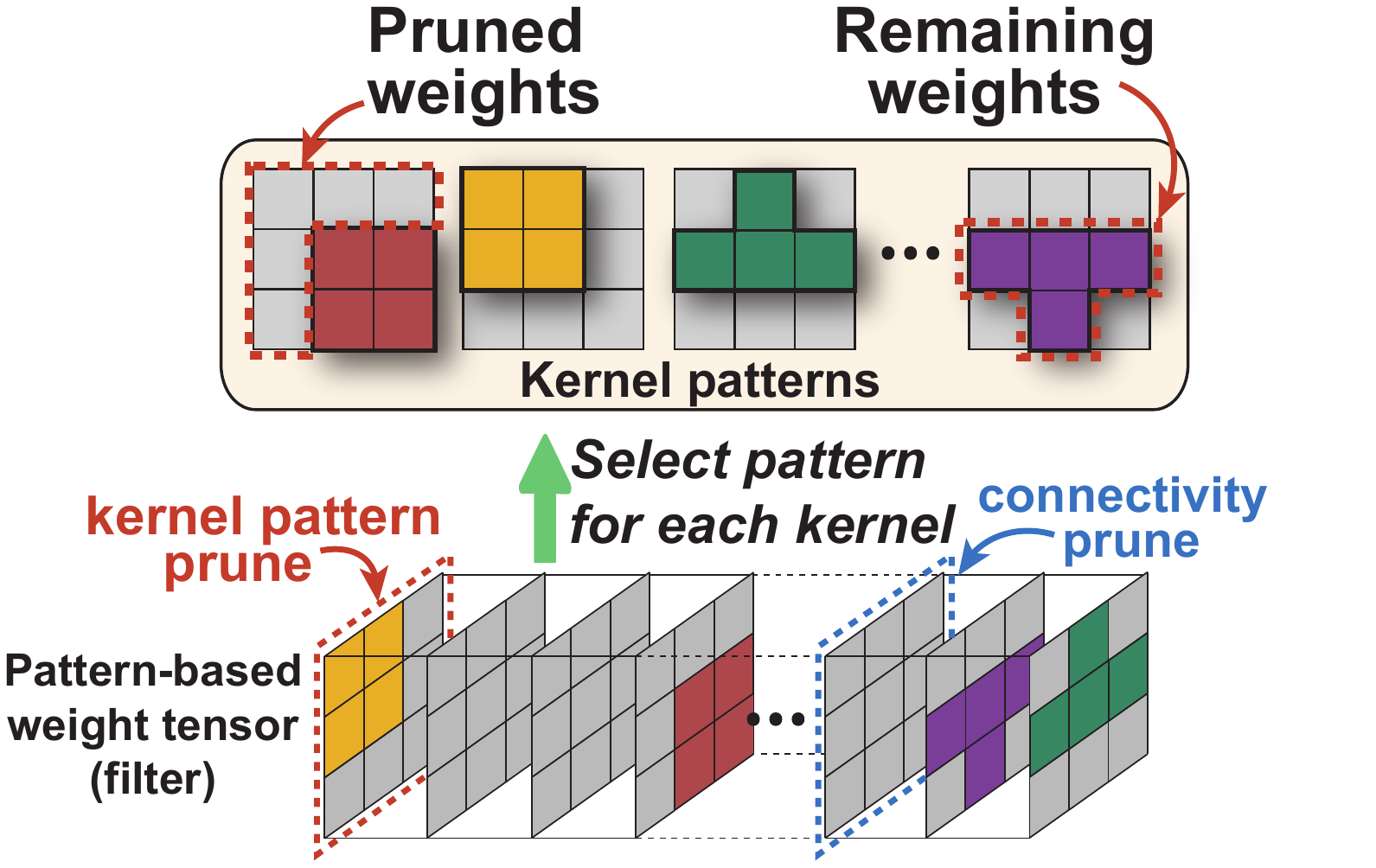}
    \caption{Illustration of pattern-based pruning.}
    \label{fig:pattern}
\end{figure}

\textbf{Pattern-based pruning} is considered as a fine-grained structured pruning scheme. It simultaneously preserves the accuracy and the hardware performance due to its proper degree of structural flexibility and structural regularity~\cite{niu2020patdnn, ma2020pconv}.
Pattern-based pruning consists of two parts, which are the kernel pattern pruning and the connectivity pruning. The kernel pattern pruning prunes a fixed number of weights in each convolution kernel, as shown in Figure~\ref{fig:pattern}.
It first analyzes the locations of remaining weights and forms a specific kernel pattern, then prunes the weights accordingly. Different kernels can apply different types of patterns but the total number of patterns is restricted to a fixed-size set. 
Connectivity pruning, as a supplementary of the kernel pattern pruning, is adopted to achieve higher overall compression rate. Connectivity pruning prunes the entire convolution kernels, which can be considered as removing the connections between certain input and output channels.
However, kernel patterns are specially designed for 3$\times$3 kernels and are not applicable to other kernel sizes. This drawback significantly restricts the use of pattern-based pruning in many scenarios. 

\subsection{Compiler-assisted DNN Acceleration on Mobile}
With the growth of mobile vision application, there is a growing need to break through the current performance limitation of mobile platforms. Both industry and academia have been putting efforts on mobile DNN execution frameworks (e.g., ~\cite{lane2016deepx,lane2015deepear,xu2018deepcache,huynh2017deepmon,yao2017deepsense,han2016mcdnn}). Among these works, TensorFlow-Lite (TFLite)~\cite{TensorFlow-Lite}, Alibaba Mobile Neural Network (MNN)~\cite{Ali-MNN}, and TVM~\cite{chen2018tvm} are three representative end-to-end DNN execution frameworks with high execution efficiency. Advanced performance optimization techniques are employed, including varied computation graph optimizations, tensor optimizations, half-float support; particularly, TVM adopts a more advanced parameters auto-tuning. However, none of those frameworks provide support for sparse (pruned) DNN models on mobile platforms\footnote{TVM considers sparsity recently for desktop processors.}. This significantly limits the performance of the DNN inference on mobile devices.

To overcome this drawback, a set of compiler-based optimizations are proposed to support sparse DNN models, significantly accelerating the end-to-end DNN inference on mobile devices. However, these optimizations are designed for fine-grained pattern-based pruning such as PatDNN~\cite{niu2020patdnn} and PCONV~\cite{ma2020pconv}, in which the specific 3$\times$3 convolution kernels in CONV layers are the main acceleration part. In addition, commonly used layers in object detection such as FC layers and 1$\times$1 CONV layers are not supported.

\section{Motivation}
\label{sec:motivation}

As mentioned above, the state-of-the-art object detection works are either accuracy-oriented using a large model size~\cite{ren2015faster,liu2016ssd,bochkovskiy2020yolov4} or speed-oriented using a lightweight model but sacrificing accuracy~\cite{sandler2018mobilenetv2,huang2018yolo,li2018tiny}. 
Therefore, all of them are hard to simultaneously satisfy the accuracy and latency demands of practical applications on mobile devices. 
As a result,
\begin{quote}
We need a solution that can achieve both high accuracy and low latency on mobile devices.
\end{quote}

\begin{figure} [t]
     \centering
     \includegraphics[width=1.0\columnwidth]{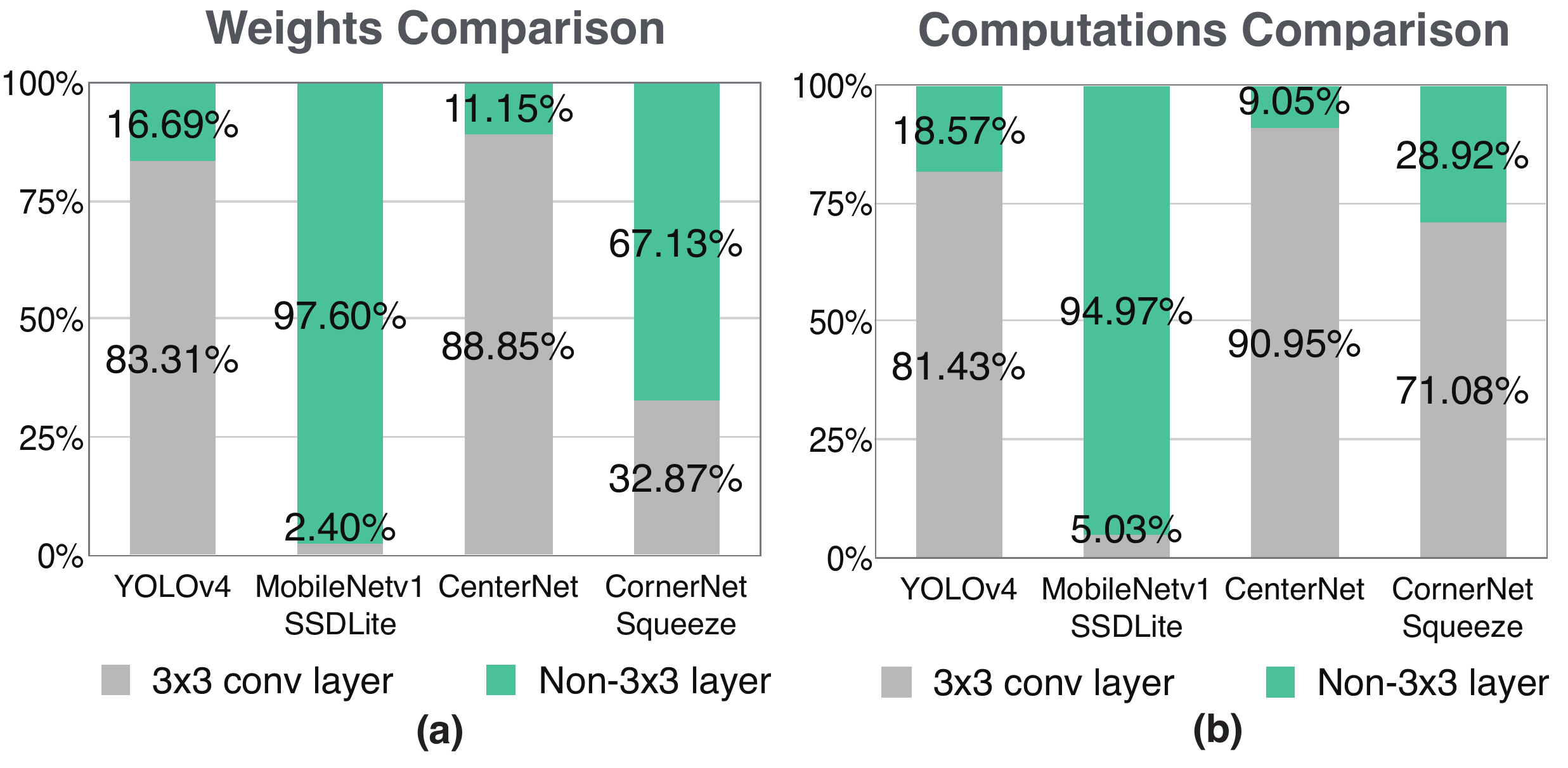}     
     \caption{Comparisons of (a) weights ratio and (b) computations ratio for 3$\times$3 convolution layers and non-3$\times$3 layers for different object detection approach.}
     \label{fig:para_ratio}
\end{figure}

While pattern-based pruning seems to be a desirable option since it strikes a balance between execution efficiency and accuracy. 
However, it is only applicable to the 3$\times$3 CONV layers and hinders its effectiveness in object detection tasks. Figure~\ref{fig:para_ratio} illustrates the comparison between $3\times3$ convolution layers and non $3\times3$ layers. We select 4 representative object detection approaches and compare the percentage of their weights and computations.  For example, in YOLOv4, one of the representative state-of-the-art networks for object detection, a considerable number of weights and amount of computations (17\% and 19\%, respectively) are contributed by non-3$\times$3 CONV layers.

Compiler-assisted DNN inference acceleration is another attractive option for low-latency DNN inference on mobile devices. It has been proved that, with the assistance of compiler optimizations, the low latency DNN inference can be achieved for image classification tasks~\cite{niu2020patdnn, ma2020pconv}.
However, such acceleration is still not sufficient enough to satisfy the low latency required by object detection tasks, since it has massive amount of weights and requires more complex computations~\cite{bochkovskiy2020yolov4, liu2016ssd}.
To this end, we raise two key design objectives:

\squishlist{}
    \item {\bf Objective 1:} We need a pruning scheme that can (i) simultaneously achieve high accuracy and leverage the underlying hardware parallelism, and (ii) be ubiquitously applied on different types of layers.
    \item {\bf Objective 2:} We need a more efficient computation method for further accelerating the DNN inference speed of object detection tasks.
\squishend{}

\section{Framework Design}
\label{sec:framework}

\subsection{Block-Punched Pruning}

\begin{figure}[t]
    \centering
    \includegraphics[width=0.94\columnwidth]{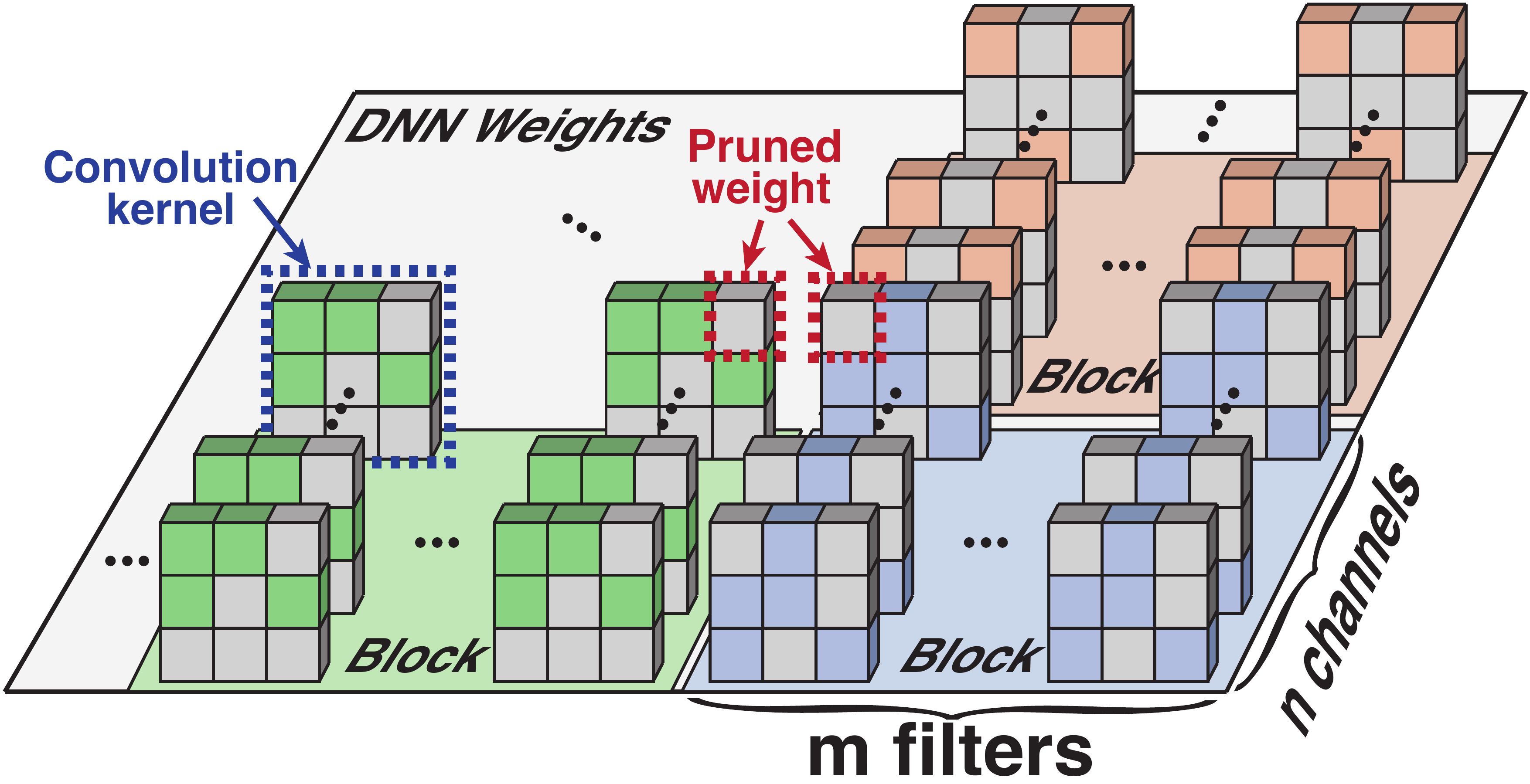}
    \caption{Illustration of block-punched pruning.}
    \label{fig:block_punched_prune}
\end{figure}
To achieve the first objective in Section~\ref{sec:motivation}, we propose a novel pruning scheme--block-punched pruning, which preserves high accuracy while achieving high hardware parallelism. In addition to the 3$\times$3 CONV layer, it can also be mapped to other types of DNN layers, such as 1$\times$1 CONV layer and FC layer. Moreover, it is particularly suitable for high-efficient DNN inference on resource-limited mobile devices.
% \hl{Let $\bm{W}_i \in \mathbb{R}^{M\times N \times K_{h} \times K_{w}}$ denote the 4-D weight tensor of the $i$-th CONV layer of CNN, where $M$ is the number of filters; $N$ is the number of input channels; $K_w$ and $K_h$ are the width and height kernels of $i$-th layer.} 
As shown in Figure~\ref{fig:block_punched_prune}, the whole DNN weights from a certain layer are divided to a number of equal-sized blocks, where each block contains the weights from $n$ consecutive channels of $m$ consecutive filters. 
In each block, we prune a group of weights at the same location of all filters while also pruning the weights at the same location of all channels. 
In other words, the weights to be pruned will punch through the same location of all filters and channels within a block. 
Note that the number of pruned weights in each block is flexible and can be different across different blocks.

\emph{\textbf{From the accuracy perspective}}, inspired by the pattern-based pruning~\cite{niu2020patdnn}, we adopt a fine-grained structured pruning strategy in block-punched pruning to increase structural the flexibility and mitigate accuracy loss. 

\emph{\textbf{From the hardware performance perspective}}, compared to the coarse-grained structured pruning, our block-punched pruning scheme is able to achieve high hardware parallelism by leveraging the appropriate block size and the help of compiler-level code generation. 
The reason is that typically the number of weights in a DNN layer is very large. 
Even when we divide the weights into blocks, the computation required by each block is still sufficient to saturate  hardware computing resources and achieve high degree of parallelism, especially on the resource-limited mobile devices.  
Moreover, our pruning scheme can better leverage the hardware parallelism from both memory and computation perspectives. First, during convolution computation, all filters in each layer share the same input. Since the same locations are pruned among all the filters within each block, these filters will skip reading the same input data, thus mitigating the memory pressure among the threads processing these filters.
Second, the restriction of pruning identical locations across channels within a block ensures that all of these channels share the same computation pattern (indices),
thus eliminating the computation divergence among the threads processing the channels within each block.

In our block-punched pruning, block size affects both the accuracy and the hardware acceleration. On the one hand, a smaller block size provides higher structural flexibility due to its finer granularity, which typically achieves higher accuracy, but at the cost of reduced speed. 
On the other hand, larger block size can better leverage the hardware parallelism to achieve higher acceleration, but it may cause more severe accuracy loss.

To determine an appropriate block size, we first determine the number of channels contained in each block by considering the computation resource of the device. For example, we use the same number of channels for each block as the length of the vector registers in the mobile CPU/GPU on a smartphone to achieve high parallelism.
If the number of channels contained in each block is less than the length of the vector registers, both the vector registers and vector computing units will be underutilized. On the contrary, increasing the number of channels will not gain extra on the performance but cause more severe accuracy drop.
Thus, the number of filters contained in each block should be determined accordingly, considering the trade-off between accuracy and hardware acceleration.

The hardware acceleration can be inferred by the inference speed, which can be obtained without the need of retraining the DNN model and is easier to derive compared with model accuracy.
Thus, a reasonable minimum required inference speed is set as the design target that needs to be satisfied. 
As long as the block size satisfies the inference speed target, we choose to keep the smallest number of filters in each block to mitigate the accuracy loss.
More detailed results will be elaborated in Section~\ref{sec:ablation}.

\subsection{Reweighted Regularization Pruning Algorithm}

In the previous weight pruning algorithms, methods such as group Lasso regularization~\cite{wen2016learning,he2017channel,liu2017learning} or Alternating Direction Methods of Multipliers (ADMM)~\cite{zhang2018systematic,ren2019admm,li2019compressing} are mainly adopted. However, it leads to either potential accuracy loss or requirement of manual compression rate tuning. Therefore, we adopt the reweighted group Lasso~\cite{candes2008enhancing} method. The basic idea is to systematically and dynamically reweight the penalties. To be more specific, the reweighted method reduces the penalties on weights with larger magnitudes, which are likely to be more critical weights, and increases the penalties on weights with smaller magnitudes.

Let $\bm{W}_i \in \mathbb{R}^{M\times N \times K_{h} \times K_{w}}$ denote the 4-D weight tensor of the $i$-th CONV layer of CNN, where $M$ is the number of filters; $N$ is the number of input channels; $K_w$ and $K_h$ are the width and height kernels of $i$-th layer. The general reweighted pruning problem is formulated as
\begin{equation} \label{prob_block}
 \underset{\bm{W},\bm{b}}{\text{minimize\quad}}
 f \big( \bm{W};\bm{b} \big)+\lambda\sum_{i=1}^{N} R(\bm{\alpha}_{i}^{(t)},\bm{W}_{i}), \\
\end{equation}
where $f \big( \bm{W};\bm{b} \big)$ represents loss function of DNN. $R(\cdot)$ is the regularization term used to generate model sparsity and the hyperparameter $\lambda$ controls the trade-off between accuracy and sparsity.
$\bm{\alpha}_{i}^{(t)}$ denotes the collection of penalty values applied on the weights $\bm{W}_i$ for layer $i$.

% Specifically, consider the kernels in a $g_m \times g_n$ block, i.e., kernels $W_i(m:m+g_m-1;n:n+g_n-1;:;:)$, weights at the same location in these kernels, i.e. $W_i(m:m+g_m-1;n:n+g_n-1;h;w)$ are determined to be pruned, where $(:;:;h;w)$ describes the same location.

Under our block-punched pruning, each $\bm{W}_i$ is divided into $K$ blocks with the same size $g_im\times g_in$, namely, $\bm W_i = [\bm W_{i1},  \bm W_{i2},...,\bm W_{iK}]$, where $\bm W_{ij} \in \mathbb R^{g_im \times g_in}$. Therefore, the regularization term is
\begin{equation}
R({\bm{\alpha}}_{i}^{(t)},\bm{W}_{i}) = \sum_{j=1}^{K}\sum_{h=1}^{g_m^{i}}\sum_{w=1}^{g_n^{i}} \big\| \alpha_{ijn}^{(t)} \circ  [\bm W_{ij}]_{h,w} \big\|_F^2,
\label{eq:column}
\end{equation}
where $\alpha^{(t)}_{ijn}$ is updated by $\alpha^{(t)}_{ijn} = \frac{1}{\|[\bm W_{ij}]^{t}_{h,w}\|_F^2 + \epsilon}$. 
% $\lambda$ is the major required hyperparameter in the reweighted method, which avoid manually deciding compression rate for each layer. 

The pruning process starts with a pre-trained DNN model. By conducting another training process using the reweighted regularization pruning algorithm, the pruned model with our block-punched constraints can be obtained.

\subsection{Mobile Acceleration with a Mobile GPU-CPU Collaborative Scheme}

\begin{figure}[!t]
    \centering
    \includegraphics[width=1.0\columnwidth]{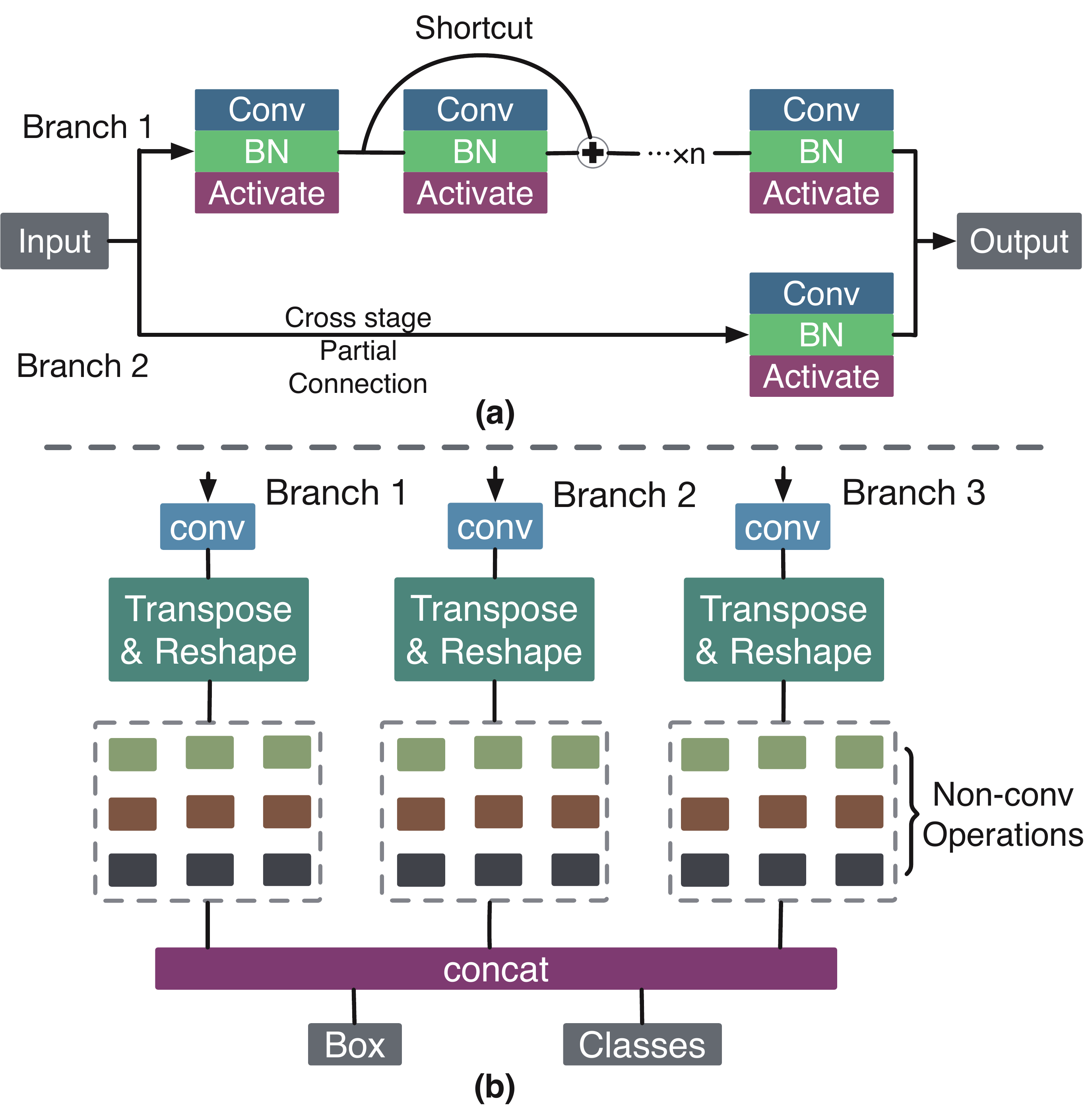}
    \caption{An illustration of the (a) cross-stage partial block and (b) non-convolutional operations in YOLO Head.}
    \label{fig:cg-co12}
\end{figure}

To achieve the second objective in Section~\ref{sec:motivation}, we propose a GPU-CPU collaborative computation scheme to improve the computational efficiency of DNNs on mobile devices.
It can be observed that the multi-branch architecture, as shown in Figure~\ref{fig:cg-co12} (a), are widely used in many state-of-the-art networks such as YOLOv4. 
Mobile devices has mobile GPU and mobile CPU, currently the DNN inference acceleration frameworks such as TFLite and MNN can only support DNN inference to be executed on either the mobile GPU or the CPU sequentially, 
which leads to a potential waste of the computation resources. The CPU is underutilized for most of the time when the GPU is computing.
Moreover, many branches have no dependencies on each other, and potentially could be computed on mobile GPU and mobile CPU concurrently to achieve higher efficiency and speed.

In our framework, we incorporate our GPU-CPU collaborative computation scheme to optimize two types of branch structures in DNNs, which are 1) the branch structure with CONV layers and 2) the branch structure with non-CONV operations. The examples of the two types of branch structures are shown in Figure~\ref{fig:cg-co12} (a) and (b). 
We do the offline device selection based on the speed before deployment. 

As we know, the GPU is suitable for high-parallelism computation, such as the convolutional computations, and it significantly outperforms the CPU in terms of speed.
Thus, \emph{\textbf{for the branch structure with CONV layers}}, such as the Cross Stage Partial (CSP) block in YOLOv4 as shown in Figure~\ref{fig:cg-co12} (a), the GPU is selected for computing the most time-consuming branch, and the problem left is to determine whether the other branches use CPU to compute concurrently or still use GPU to compute sequentially. 

In Figure \ref{fig:cg-co12} (a), we name the GPU computing time in branch 1 and branch 2 as $t_{g1}, t_{g2}$, CPU computing time as $t_{c1}, t_{c2}$, and data copying time as $\tau$.
We execute the most time-consuming branch 1 in GPU and make a decision for branch 2. 
When using CPU for parallel computing, we also need to add the data copying time $\tau$. 
The desired GPU-CPU parallel computing time $T_{par}$ depends on the maximum time cost of the branch 1 and branch 2:
$$T_{par} = max\{t_{g1} \, ,\, t_{c2}+\tau  \}$$

The GPU-only serial computing time $T_{ser}$ is the summation of computing time $t_{g1}+t_{g2}$ of two branches:
$$T_{ser}= t_{g1}+t_{g2}$$

Based on the minimum of GPU-CPU parallel computing time $T_{par}$ and GPU-only computing time $T_{ser}$, we can select the optimal executing device for branch 2.
Note that the determination of execution devices for each branch structure in YOLOv4 is independent to other branch structures. Thus, the execution devices for all branch structures in the network can be solved by greedy algorithm~\cite{cormen2009introduction}.

On the other hand, limited by the power and area, mobile GPUs usually have lower performance. For the less computational intensive operations, such as point-wise add operation and point-wise multiplication operation, mobile CPU performs similar or even faster speed compared with mobile GPU.
Therefore,  \emph{\textbf{for the branch structures with non-CONV operations}}, either of CPU or GPU can be used for each branch depending on total computation time.

Take the three final output YOLO head structures in YOLOv4 as an example, as shown in Figure~\ref{fig:cg-co12} (b). 
After transposing and reshaping the output from the last CONV layer in each branch, we still need several non-CONV operations to get the final output. We measure the total GPU and CPU execution times for the non-CONV operations in each branch and denote them as $t_{g0}, t_{g1}, t_{g2}$ and $t_{c0}, t_{c1}, t_{c2}$ respectively. 
The $T_{total}$ denotes the total computing time for all three branches.

Now we have eight possible combinations of device selections for the three branches. For example, if first two branches use CPU and the third branch uses GPU, the total computing time will be
$$T_{total} = max\{t_{c0}+t_{c1} \, ,\, t_{g2} \}$$
Note that the final output has to be moved to CPU sooner or later, so we do not count the data copying time into the total computation time.
As a result, we select the combination that has the minimum total computation time as our desired computation scheme.
Putting all together, our proposed GPU-CPU collaborative scheme can effectively increase the hardware utilization and improve the DNN inference speed.

\subsection{Compiler-assisted Acceleration}

Inspired by PatDNN~\cite{niu2020patdnn}, YOLObile relies on several advanced compiler-assisted optimizations that are enabled by our newly designed block-punched pruning to further improve the inference performance. 
We summarize them here briefly due to the space constraints. First, YOLObile stores the model weights compactly by leveraging the pruning information (the block and punched pattern) that can further compress the index arrays comparing to the well-known Compressed Sparse Row format. Second, YOLObile reorders blocks to improve memory and computation regularity, and to eliminate unnecessary memory access. Moreover, YOLObile employs a highly parallel auto-tuning model to find the best execution configuration parameters. 
YOLObile generates both CPU and GPU codes for each layer, and calls the right one according to our GPU-CPU collaborative scheme during the actual inference process.

\section{Evaluation}
\label{sec:evaluation}

In this section we evaluate our proposed YOLObile framework on mobile devices in terms of accuracy and inference speed, compared with other state-of-the-art frameworks. Additionally, ablation study on different pruning schemes and configurations are provided. 

\textbf{Experimental Setup} Our models are trained on a server with eight NVIDIA RTX 2080Ti GPUs. The training methods are implemented using PyTorch API.
We evaluate our framework on an off-the-shelf Samsung Galaxy S20 smartphone, which has a Qualcomm Snapdragon 865 Octa-core CPU and a Qualcomm Adreno 650 GPU.
Each test runs on 50 different input frames (images), with the average speed results reported. 
Our YOLObile is derived based on YOLOv4, with 320$\times$320 input size, and train on MS COCO dataset~\cite{lin2014microsoft}. 
We denote mAP as the Average Precision under IoU 0.5 threshold and AP@[.5:.95] as the  Average Precision under IoU from 0.5 to 0.95.
% We do not test our result on Pascal VOC, for MS COCO has more accurate labels and higher threshold of IOU.
Note that our compiler achieves much higher speed for object detection approaches compared with existing compiler-assisted frameworks, such as TFLite. More comparisons are presented in supplementary materials.

\subsection{Evaluation of block-punched pruning} 
% \textbf{Compression Rate \& Accuracy.} 
We first evaluate the accuracy and compression rate of our proposed block-punched pruning in YOLObile framework. As mentioned above, block size affects both accuracy and hardware acceleration performance. We adopt 8$\times$4 as our block size, i.e. 4 consecutive channels of 8 consecutive filters. The details of the impact of different block sizes are discussed in ablation study \ref{sec:ablation}.   
The original YOLOv4 model contains 64.36M weights and requires 35.8G floating-point operations (FLOPs). As shown in Table \ref{tab:se5tb1}, by applying our block-punched pruning, we achieve the compression rate up to 14$\times$ (in weights) with 49 mAP. The weight number decreases to 4.59M and FLOPs is reduced to 3.59G. 
With 92.87\% weights and 88.97\% FLOPs reduced, our model still maintains a decent accuracy, with only 8.3 mAP loss.

% \begin{table}[!t]
%     \centering
%     % \tiny
%     \footnotesize 
%     % \small
%     \setlength{\tabcolsep} {1.2mm}
%     \begin{tabular}{|c|c|c|c|c|c|c|}
%          \hline
%          \bfseries \#Weights & \bfseries \tabincell{c} {\#Weights\\ Comp. Rate} & \bfseries \#FLOPs & \bfseries \tabincell{c} {\#FLOPs\\ Comp. Rate} &$\mathbf{mAP_{0.5}}$ &$\mathbf{AP_{0.5:0.95}}$ & \bfseries FPS \\
%          \hline
%          64.36M &   1$\times$       &   35.8G   &	1$\times$   &	57.3    &   38.2    &	3.5 \\\hline
%          16.11M	&   3.99$\times$    &	10.48G  &	3.41$\times$&	55.1    &	36.5	&   7.3 \\
%          8.04M  &	8.09$\times$    &	6.33G   &	5.65$\times$&   51.4    &	33.3	&   11.5 \\
%          6.37M  &	10.1$\times$	&   5.48G   &	6.53$\times$&	50.9    &	32.8    &	13  \\
%          4.59M  &	14.02$\times$   &	3.95G   &	9.06$\times$&	49      &   31.9	&   17 \\
%          \hline
%     \end{tabular}
%         \caption{Accuracy and speed of under different compression rates.}
%     \label{tab:se5tb1}
% \end{table}

\begin{table}[!t]
    \centering
    % \tiny
    % \footnotesize 
    \small
    \setlength{\tabcolsep} {1.2mm}
    \begin{tabular}{cccccc}
         \toprule
         \bfseries \#Weights & \bfseries \tabincell{c} {\#Weights\\ Comp. Rate} & \bfseries \#FLOPs  &\bfseries mAP &\bfseries AP@[.5:.95] & \bfseries FPS \\
         \midrule
         64.36M &   1$\times$       &   35.8G   &	57.3    &   38.2    &	3.5 \\\midrule
         16.11M	&   3.99$\times$    &	10.48G  &	55.1    &	36.5	&   7.3 \\
         8.04M  &	8.09$\times$    &	6.33G   &   51.4    &	33.3	&   11.5 \\
         6.37M  &	10.1$\times$	&   5.48G   &	50.9    &	32.8    &	13  \\
         4.59M  &	14.02$\times$   &	3.95G   &	49      &   31.9	&   17 \\
         \bottomrule
    \end{tabular}
        \caption{Accuracy and speed under different compression rates.}
    \label{tab:se5tb1}
\end{table}
\begin{table*}[!htbp]
    \centering
    % \tiny
    % \footnotesize 
    \small
    \setlength{\tabcolsep} {1.8mm}
    \begin{threeparttable} [t]
    \begin{tabular}{cccccccc}
         \toprule
         \bfseries Approach & \bfseries Input Size & \bfseries backbone & \bfseries \#Weights & \bfseries \#FLOPs &\bfseries mAP &\bfseries AP@[.5:.95]  & \bfseries FPS \\
         \toprule
         CenterNet-DLA (\citeauthor{duan2019centernet})              &	512 &	DLA34	    &   16.9M   &	52.58G  &	57.1    &	39.2    &	1.9 \\
         CornerNet-Squeeze (\citeauthor{law2019cornernet})           &  511 &	-   	    &   31.77M  &	150.15G &	-   	&   34.4    &	0.3 \\
         SSD (\citeauthor{liu2016ssd})                               &	300 &	VGG16       &	26.29M  &	62.8G   &	43.1    &	25.1    &	4.2 \\
         MobileNetv1-SSDLite (\citeauthor{sandler2018mobilenetv2})   &	300 &	MobileNetv1 &	4.31M   &	2.30G   &	-	    &   22.2    &	49  \\
         MobileNetv2-SSDLite (\citeauthor{sandler2018mobilenetv2})   &	300 &	MobileNetv2	&   3.38M   &	1.36G   & 	-       &	22.1    &	41  \\
         Tiny-DSOD (\citeauthor{li2018tiny})                         &	300 &	-	        &   1.15M   &	1.12G	&   40.4    &	23.2	&   -   \\
         YOLOv4 (\citeauthor{bochkovskiy2020yolov4})                 &  320 &	CSPDarknet53&	64.36M  &	35.5G	&   57.3    &   38.2    &   3.5 \\
         YOLO-Lite (\citeauthor{huang2018yolo})                      &	224 &	-           &	0.6M	&   1.0G    &	 -	    &   12.26	&   36  \\
         YOLOv3-tiny (\citeauthor{redmon2018yolov3})                 &	320 &	Tiny Darknet&	8.85M   &	3.3G    &	29      &	14      &	14  \\
         YOLOv4-tiny (\citeauthor{bochkovskiy2020yolov4})	         &  320 &	Tiny Darknet&	6.06M   &	4.11G	&   40.2	&   -       &	11  \\
         \textbf{YOLObile (GPU only)}	                             &  320	&   CSPDarknet53&	4.59M	&   3.95G	&   \textbf{49}	    &   \textbf{31.6}	&   \textbf{17}\\
         \textbf{YOLObile (GPU\&CPU)}                                &  320	&   CSPDarknet53&	4.59M	&   3.95G	&   \textbf{49}	    &   \textbf{31.6}	&   \textbf{19.1}  \\
         \bottomrule
    \end{tabular}
    \end{threeparttable}
        \caption{Accuracy (mAP) and speed (FPS) comparison with other object detection approaches.}
    \label{tab:se5tb2}
\end{table*}

\subsection{Evaluation of YOLObile framework}

\begin{figure} [t]
     \centering
     \includegraphics[width=0.9\columnwidth]{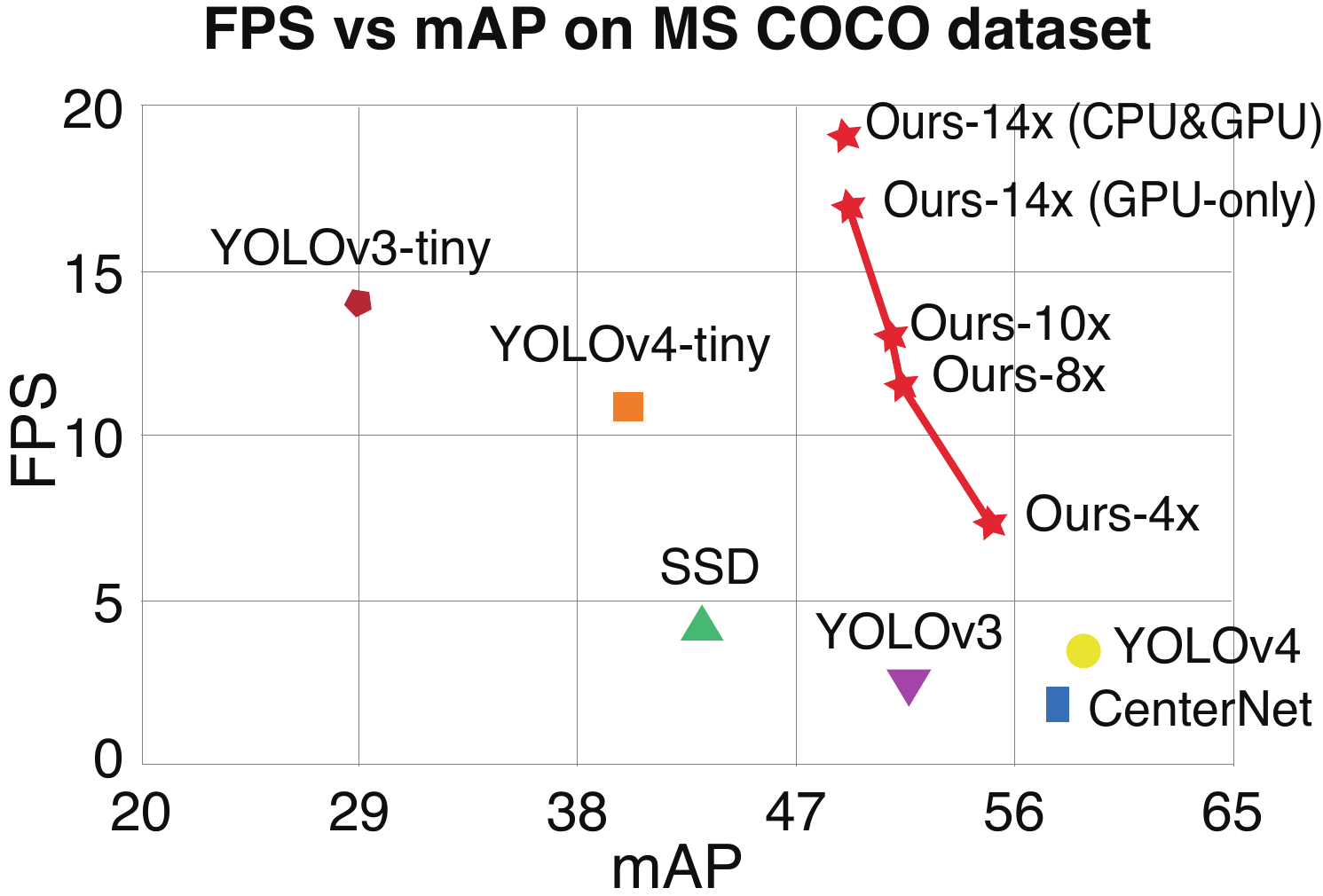}     
     \caption{The accuracy (mAP) and speed (FPS) comparison of YOLObile under different compression rate and different approaches.}
     \label{fig:se5fpsvsap}
\end{figure}

To validate the effectiveness of our framework, we compare our YOLObile with several representative works. 
To make fair comparisons, all the results (including the object detection approaches from the reference works) are evaluated under our compiler optimizations.
As shown in Table~\ref{tab:se5tb2}, under a similar number of computations and even having a smaller model size, YOLObile consistently outperforms YOLOv3-tiny and YOLOv4-tiny in terms of mAP and FPS. 
This indicates our proposed block-punched pruning is a more desired method to achieve a smaller model size while maintaining the mAP compared to training a small model from scratch.

YOLObile achieves even higher mAP than the full-size one-stage detector SSD but lower mAP than YOLOv4 and CenterNet. However, the inference speed of YOLObile is much faster than SSD, YOLOv4 and CenterNet (4.5$\times$, 5.5$\times$ and 10$\times$ respectively). 
Comparing with the lightweight detectors such as YOLO-Lite and MobileNetv2-SSDLite, YOLObile has lower FPS but much higher mAP.
Figure~\ref{fig:se5fpsvsap} demonstrates the mAP and FPS of YOLObile under different compression rates and the results are compared with representative reference works.
Our YOLObile lies in top right of the figure, and outperforms YOLOv3, SSD, YOLOv3-tiny and YOLOv4-tiny in both accuracy and speed.
Unlike the lightweight approaches, which simply trade the mAP for FPS, YOLObile provides a Pareto-Optimal trade-off solution that maintains both the mAP and FPS.

We also evaluate the performance of our GPU-CPU collaborative computation scheme.  As shown in Table~\ref{tab:se5tb2}, comparing to the GPU-only execution, our GPU-CPU collaborative computation scheme effectively accelerates the inference speed and improves FPS.

\subsection{Ablation Study}
\label{sec:ablation}

\textbf{Ablation study on pruning scheme.} In this section, we conduct experiments on YOLOv4 under different pruning schemes. 
% report the accuracy and speed. 
Table \ref{tab:se7prunscheme} shows the comparison of different pruning scheme results under 8$\times$ compression rate. 
Unstructured pruning scheme achieves the highest mAP because of its flexibility. However, the inference speed in FPS is only 6.4 due to underutilized hardware parallelism.
Structured pruning (filter pruning) shows high inference speed, but with severe accuracy drop. 
Compared with structured pruning and unstructured pruning, our block-punched pruning scheme achieves both high accuracy and fast inference speed.

\begin{table}[!t]
    \setlength\tabcolsep{4pt}
    % \footnotesize
    \small
    \centering
    \begin{tabular}{ccccc}
    \toprule
    \bfseries \tabincell{c} {Pruning\\ Scheme}   & \bfseries \#Weights & \bfseries \tabincell{c} {\#Weights\\ Comp. Rate} &\bfseries mAP & \bfseries FPS \\
    \midrule
    Not Prune   &	64.36M  &	1$\times$   &	57.3    &	3.5 \\
    Unstructured&	8.04M	&   8.09$\times$&	53.9	&   6.4 \\
    Structured  &	8.04M	&   8.09$\times$&	38.6	&   13  \\
    Ours        &	8.04M   &	8.09$\times$&	51.4    &	11.5\\
    \bottomrule
    \end{tabular}
        \caption{Comparison of different pruning schemes.}
    \label{tab:se7prunscheme}
\end{table}

\begin{figure}[!t]
    \centering
    \includegraphics[width=1.0\columnwidth]{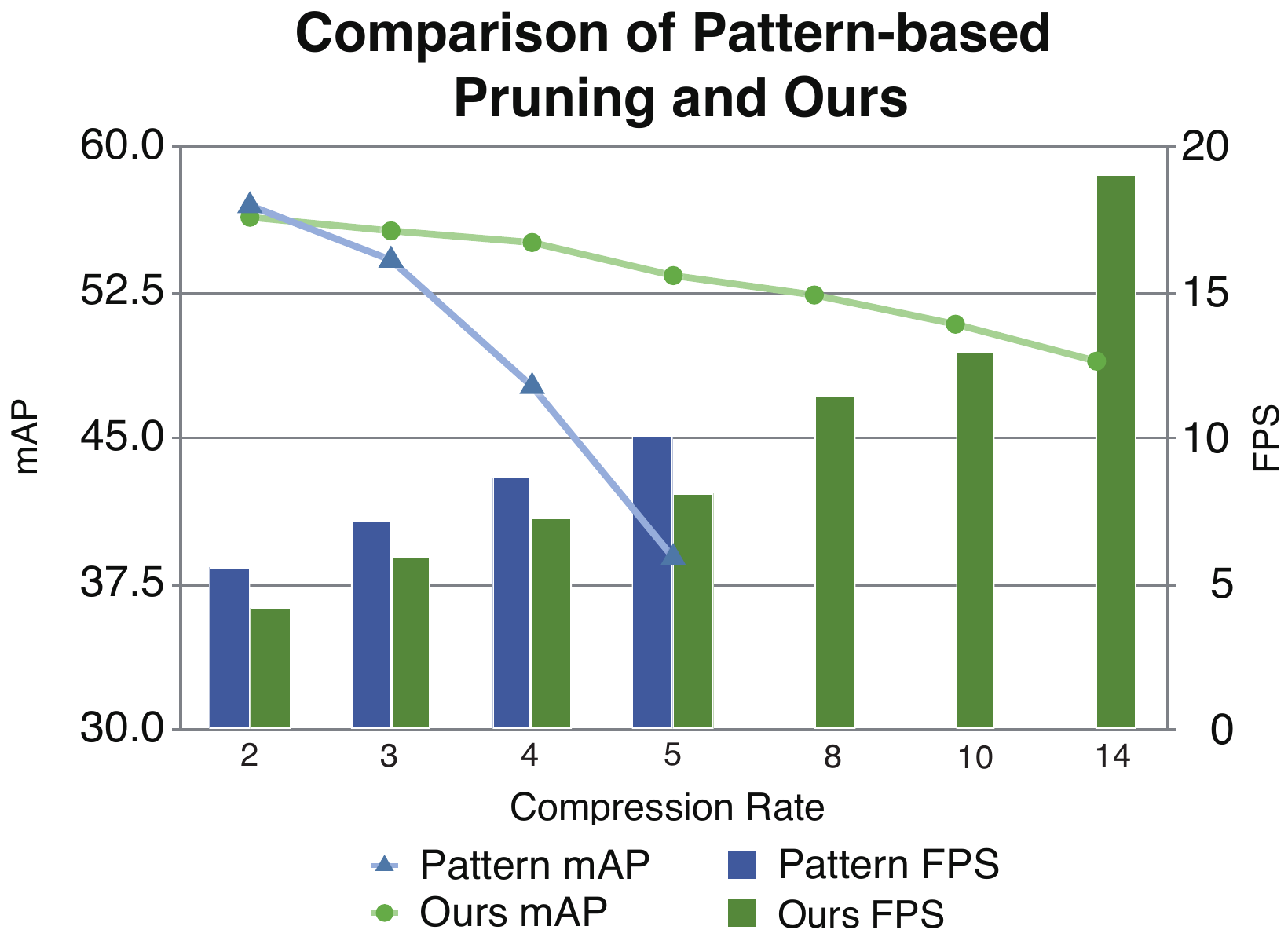}
    \caption {mAP and FPS comparison of pattern-based pruning and ours.}
    \label{fig:se7_pat}
\end{figure}

\textbf{Ablation Study on pattern-based pruning.}
Despite the promising performance that pattern-based pruning can achieve, it has the restriction of kernel size to be 3, while in non-3$\times$3 layers it cannot be applied. Unfortunately, in object detection approaches such as YOLOv4, the ratio of 3$\times$3 CONV layers is 83.31\% in total weights, which limits the highest compression rate of pattern-based pruning to 5.99$\times$. Therefore, we compare pattern-based pruning and our block-punched pruning scheme under compression rate of 2$\times$, 3$\times$, 4$\times$, 5$\times$, respectively. Additionally, we demonstrate the result under our block-punched pruning scheme with 8$\times$, 10$\times$ 
and 14 $\times$ compression rate. In Figure~\ref{fig:se7_pat}, we plot the mAP curve and FPS bar under different compression rate. 
We can see when the compression rate is below 3$\times$ pattern-based pruning has higher accuracy as it is more flexible. When the compression rate increases, for pattern-based pruning, we have to prune more weights in each 3$\times$3 CONV layer, because non-3$\times$3 layers can not be pruned. 
The extremely high layer-wise prune ratio results in a sharp drop down of the curve. 
Pattern-based pruning has higher inference speed compared to our block-punched pruning. However, the compression rate ceiling limits the inference speed. Ours scheme can reach higher speed when compression rate increases, and ours do not suffer from sharply accuracy drop down.

\begin{figure}[!t]
    \centering
    \includegraphics[width=0.9\columnwidth]{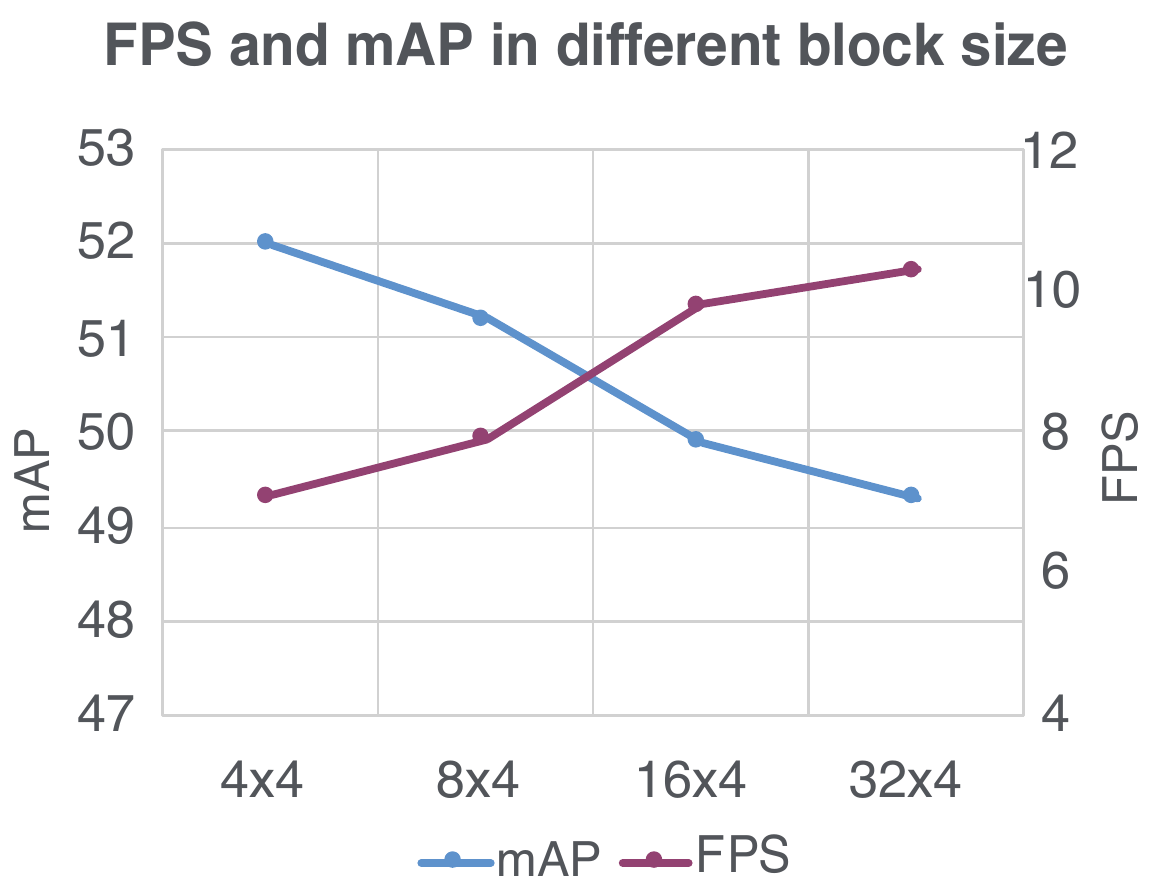}
    \caption {Accuracy (mAP) and speed (FPS) of different block size pruning results.}
    \label{fig:se7_blksize}
\end{figure}

\textbf{Ablation study on block size.} We conduct experiments on four different block sizes using our block-punched pruning scheme to check the impact of block size on results.
To achieve high hardware parallelism, the number of channels in each block is fixed to 4, which is the same as the length of GPU and CPU's vector registers.
The accuracy and speeds are evaluated under different numbers of filters in each block.
As shown in Figure~\ref{fig:se7_blksize}, larger block size can better leverage the hardware parallelism compared with smaller block size and achieves higher inference speed. However, it leads to accuracy loss due to its coarse pruning granularity. 
Smaller block size can achieve higher accuracy but sacrifice the inference speed.
According to the results, we consider 8$\times$4 (4 consecutive channels of 8 consecutive filters) as a desired block size on mobile devices, which strikes a good balance between both the accuracy and the speed.

\textbf{Ablation study on layer-wise compression rate between different kernel size.}
YOLOv4 contains only 3$\times$3 kernel size and 1$\times$1 kernel size in CONV layers. 
We believe these 2 types of CONV layers have different levels of sensitivity in pruning process, therefore we conduct 2 groups of experiments. Under the same number of FLOPs, we evenly prune all the layers in one group. And in another group, the compression rate of 3$\times$3 CONV layers is 1.15$\times$ higher than in 1$\times$1 CONV layers. 
As shown in Table \ref{tab:se7even}, the evenly pruned model exhibits lower accuracy and lower inference speed than the unevenly pruned model. Since 3$\times$3 CONV layers contributes 81.4\% of the total FLOPs, it can be concluded that compression rates in these layers illustrate a higher impact on the overall performance.

\begin{table}[!t]
    % \setlength\tabcolsep{3pt}
    % \footnotesize
    \small
    \centering
    \begin{threeparttable}
    \begin{tabular}{ccccc}
    \toprule
    \bfseries Pruning method & \bfseries \#Weights & \bfseries \#FLOPs & \bfseries mAP & \bfseries FPS \\
    \midrule
    Evenly Prune             &	10.38M  &	6.2G    &	50.5    &	8 \\
    % \hline
    Unevenly Prune  &	8.04M   &   6.2G    &	51.4	&   11.5 \\
    \bottomrule
    \end{tabular}
    % \begin{tablenotes}
    %     \scriptsize
    %     \item[*] We use the layerwise compression rate of $3\times3$ CONV:$1\times1$ CONV = 1.15:1 in unevenly prune.
    %   \end{tablenotes}
    \end{threeparttable}
    \caption{Evenly Prune vs Unevenly Prune.}
    \label{tab:se7even}
\end{table}
\section{Conclusion}

In this work, we propose YOLObile, a real-time object detection framework on mobile devices via compression-compilation co-design. 
A novel pruning scheme—-block-punched pruning is also proposed, designed for CONV layers with \emph{any} kernel size as well as fully-connected (FC) layers.
To improve the computational efficiency of DNNs on mobile devices, the proposed YOLObile also features a GPU-CPU collaborative computation scheme in addition to our proposed compiler optimizations.
The evaluation demonstrates that our YOLObile framework exhibits high accuracy while achieving high hardware parallelism.

\bibliography{reference}

\end{document}